\documentclass[10pt,journal,compsoc]{IEEEtran}
\usepackage{amsmath,amsfonts}
\usepackage{algorithmic}
\usepackage{algorithm}
\usepackage{array}
\usepackage[caption=false,font=normalsize,labelfont=sf,textfont=sf]{subfig}
\usepackage{textcomp}
\usepackage{stfloats}
\usepackage{url}
\usepackage{verbatim}
\usepackage{graphicx}
\usepackage{cite}
\usepackage{booktabs}
\usepackage{lipsum}
\usepackage{mwe}
\usepackage{ccicons}
\usepackage{mathptmx}
\usepackage{amsmath}
\usepackage{multirow}
\usepackage{amssymb}
\usepackage{placeins}
\usepackage[listings]{tcolorbox}
\tcbuselibrary{breakable, skins}
\usepackage{inconsolata}
\usepackage{cleveref}

\newtcblisting{mylisting}[2][]{
    breakable,
    colback=white,
    colframe=black,
    sharp corners,
    boxrule=0.4pt,
    left=2mm, right=2mm, top=2mm, bottom=2mm,
    listing only,
    title={#2},
    label={#1},
    colbacktitle=white,
    coltitle=black,
    title style={
        colframe=black,
        boxrule=0.4pt,
        sharp corners,
        left=1mm,
        right=1mm,
        top=0.5mm,
        bottom=0.5mm,
        titlerule=0.4pt,
        titlerule color=black
    },
    listing options={
        basicstyle=\small\ttfamily,
        breaklines=true,
        breakindent=0pt
    }
}
\newcommand{\eg}{\textit{e}.\textit{g}.}

\usepackage{caption}

\begin{document}

\title{Scalable Multi-Human Cooperative Manipulation via Commander-Executor Coordination}

\author{Buzhen Huang, Xingchen Wu, Yuanwang Yang, Chenyi Xie and Kun Li$^*$
\thanks{Buzhen Huang, Xingchen Wu, Yuanwang Yang, Chenyi Xie, and Kun Li are with Tianjin University. }
\thanks{$^*$ Corresponding author: Kun Li. E-mail: lik@tju.edu.cn.}
}

\IEEEpubid{0000--0000/00\$00.00~\copyright~2021 IEEE}

\maketitle

\begin{abstract}
Multi-human cooperative object manipulation is a highly challenging topic in humanoid control. While existing methods can handle simple collaborative carrying tasks, they struggle with role-dependent, long-horizon, and sequential cooperative manipulation. To address this issue, we propose a hierarchical commander-executor framework that enables multiple agents to accomplish complex collaborative tasks in distinct ways. Specifically, a commander combines instructions and scene information for joint semantic and geometric planning. It then decomposes the task into role-specific subtasks and generates shared object-centric coordination plans, specifying execution timing, target positions, and manipulation actions for each agent. Conditioned on the commander's high-level strategy and each agent's local observations, the executor learns a low-level physics-based policy through a solo-to-cooperative curriculum to realize coordinated human-object interactions. The planned object trajectory provides a shared coordination target, while root positions and grasping instructions specify role-dependent execution cues. We evaluate our method on a variety of object manipulation tasks, demonstrating its ability to execute long-horizon tasks that combine different types of collaboration and to scale from two-agent to multi-agent cooperation. Extensive experiments further show that our approach achieves higher success rates compared to state-of-the-art methods. Code will be made publicly available at \url{https://github.com/TJU-3DV/Co-Manipulation}.
\end{abstract}


\begin{figure*}[h]
  \centering
  \includegraphics[width=\linewidth]{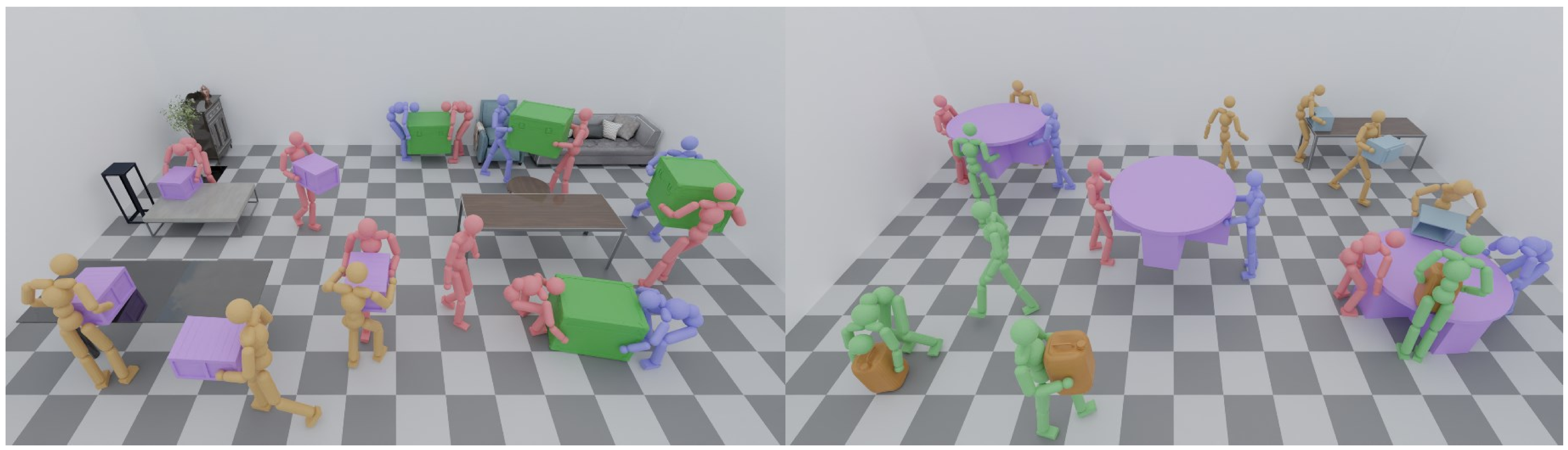}
  \vspace{-7mm}
  \caption{Given a scene mesh together with a task instruction, our method plans a manipulation strategy using task semantics and scene geometry, and controls humanoids to produce coordinated humanoid-object interaction motions for long-horizon collaborative manipulation (\eg, handover and cooperative carrying) across diverse object types.}
  \vspace{-3mm}
  \label{fig:teaser}
\end{figure*}

\section{Introduction}\label{sec:intro}

\IEEEPARstart{L}{ong-horizon} collaborative object manipulation among multiple ag-ents is a crucial capability for embodied humanoid systems in complex scenarios. It requires a diverse set of skills, including cooperative transportation, object handover, and collision-free coordination in cluttered or constrained environments. However, enabling such collaboration remains challenging, as agents must plan long-horizon manipulation strategies while adapting their actions to one another throughout execution.

Conventional kinematics-based methods~\cite{xu2026stability,liang2024intergen,li2023object} often produce physically implausible artifacts such as penetration and sliding, limiting their applicability to embodied interaction tasks. Recently, dynamics-based methods have been able to produce physically plausible motion, but most existing works remain limited to single-agent object interaction~\cite{pan2025tokenhsi,xu2025intermimic,xu2026interprior}. Only a few works have attempted to model multi-agent object interaction. However, these methods either rely on simplified object assumptions~\cite{gao2024coohoi} for collaboration or restrict interactions to specific contact patterns~\cite{lionar2026teamhoi}. Moreover, existing approaches mainly focus on cooperative carrying, limiting their ability to handle diverse collaborative behaviors such as handover and long-horizon interactions in complex scenarios.

We observe that in real-world collaborative tasks,\IEEEpubidadjcol a commander plays a key role in ensuring successful task execution and significantly enhancing team efficiency. In a long-horizon complex collaboration task, the commander understands the global objective, decomposes it into coordinated subtasks, assigns responsibilities to different roles, and provides a high-level strategy for collaboration. The performers execute the planner's scheme by following their assigned roles, coordinating with one another through shared task references, and progressively adapting their behaviors from individual execution to cooperative interaction. Based on this observation, we propose a commander-executor framework that first generates task plans by jointly considering task semantics and scene geometry, and then trains a collaboration-aware control policy for executing diverse interaction tasks in complex scenes.

Specifically, the commander first represents the scene configurations as scene graph and uses LLMs~\cite{liu2024deepseek} with an instruction to reason about the semantic relations among agents, objects, and goals. Since semantic reasoning alone may produce plans that are difficult to execute physically, we further incorporate scene geometry to derive feasible coordination cues, including obstacle-aware human and object trajectories and role-specific execution cues. These cues serve as a shared task reference that specifies not only the final goal, but also how different agents should coordinate over time. Conditioned on this high-level strategy and each agent's local observations, the executor learns a control policy with sparse guidance rather than full-body motion references, reducing the dependence on scarce multi-agent interaction data. To improve stability and scalability, we train the single-agent policy with augmented objects, contacts, and trajectories, enabling it to learn robust multi-phase manipulation skills such as approaching, grasping, and carrying. Since the single-agent policy shares the same object-centric strategy representation as in collaborative tasks, the learned policy can be efficiently transferred to multi-agent cooperation without relying on paired multi-agent interaction data. As a result, given a scene mesh and object-level task goals, our method can generate feasible manipulation strategies and control multiple humanoids to perform coordinated long-horizon interactions, such as handover and cooperative carrying, across diverse object types.
The main contributions of this work are summarized as follows:
\begin{itemize}
    \item We propose a scalable Commander--Executor framework for long-horizon collaborative object manipulation that supports diverse collaboration types, different object categories, and arbitrary numbers of collaborating agents.
    \item We introduce a semantic- and geometry-aware Commander that jointly considers task semantics and scene geometry to generate feasible coordination strategies, supporting reasonable multi-agent planning and successful execution in complex scenes.
    \item We propose a solo-to-cooperative learning scheme that leverages data augmentation in single-agent training to acquire robust manipulation skills, and efficiently transfers the policy to multi-agent collaboration, where all agents are controlled by a single shared policy.
\end{itemize}
\section{Related Work}\label{sec:Related}

\noindent\textbf{Human-human-objection interaction.}
Human-Human-Object Interaction requires modeling both human-object interaction and multi-person collaboration. Recent datasets such as Core4D~\cite{liu2025core4d} and InterVLA~\cite{xu2025perceiving} have advanced this direction. Recent generative methods have explored different ways to model collaborative human-object-human interactions. COLLAGE \cite{daiya2025collage} combines large language models with hierarchical latent diffusion to generate collaborative interactions. OnlineHOI~\cite{ji2025onlinehoi} introduces Mamba with a memory-augmented network for online multi-person-object motion generation from historical states. SyncDiff~\cite{he2025syncdiff} improves coordination through synchronized motion diffusion, while HHOI~\cite{na2025learning} composes human-object and human-human interaction priors to synthesize human-human-object interactions. Centering on object trajectory, \cite{daiya2026sketch2colab,xu2026stability} use a flow matching network to generate interactive motions between two persons and object. However, these methods do not fully consider the physical constraints. Beyond generative approaches, some methods introduce physical constraints or explicit interaction modeling. QP-based methods~\cite{vaillant2016multi} require predefined contact points and often produce less natural motions, while IG-based methods \cite{zhang2023simulation} rely on reference motion imitation and lack generalizability. Recent physics-based methods such as CooHOI~\cite{gao2024coohoi} and TeamHOI~\cite{lionar2026teamhoi} improve multi-person carrying with object dynamics and inter-agent coordination, but remain limited to simple cooperative carrying and cannot support diverse collaboration patterns. In contrast, our method combines task- and geometry-aware strategy planning with physics-based policy execution, enabling diverse long-horizon collaborative manipulation across different objects, collaboration types, and numbers of agents.

\noindent\textbf{Kinematics-based motion generation.}
Kinematics-based methods focus on generating visually smooth and coherent human motions, often without explicit modeling of physical laws such as gravity or contact forces. Approaches such as  ~\cite{zhang2022motiondiffuse,petrovich22temos,Guo_2022_CVPR,humantomato,liang2024intergen,tevet2023human,huang2024stablemofusion,shan2024opendomaintextdrivensynthesis,ota2025pino,chang2025largescalemulticharacterinteractionsynthesis} generate human motion from textual descriptions or vision clips, yet they are generally limited to single-person or person-to-person interactions, without incorporating objects. In contrast, methods like ~\cite{razali2023action,li2023object,ghosh2023imos,li2024controllable,li2024task,zeng2025chainhoi,wu2025human,wang2026unleashing,peng2025hoi} extend motion generation to human–object interactions, producing motions conditioned on text or specific action instructions. However, these methods remain largely constrain-ed to single-person scenarios and, due to the lack of physical constraints, often suffer from artifacts such as foot sliding, penetration, and physically implausible motions. Our method addresses these issues by leveraging physics simulation, enabling physically consistent motion generation for two-person object interactions.

\noindent\textbf{Physics-based motion generation.}
Physics-based motion generation methods typically train control policies within a physics simulator using reinforcement learning to produce physically plausible behaviors. Early works \cite{peng2018deepmimic,wang2023physhoi,Luo2023PerpetualHC,Wang_2025_CVPR,xu2025intermimic} primarily focused on motion imitation, relying on carefully designed motion tracking reward functions to drive policies toward accurately reproducing given motion capture sequences. However, the capabilities of imitation learning are inherently limited by the available reference motions. To enable characters to address open-domain, task-oriented problems, recent research has shifted toward reference-free generation paradigms. Representative methods based on Motion Priors~\cite{peng2021amp,mu2025smp,zhang2025physics} replace tracking of individual reference motions with a discriminator that provides style-based rewards. Combined with task-specific objective rewards, \cite{hassan2023synthesizing,pan2025tokenhsi}successfully accomplish targeted tasks in a human-like manner. Other methods like~\cite{tessler2024maskedmimic,xu2026interprior,tessler2025maskedmanipulator} use knowledge distillation, where a teacher policy for full-dimensional tracking is distilled into a student policy for partial tracking, thereby enabling diverse human-object interactions. Similar to our work, recent studies ~\cite{xie2023hierarchical,pan2024synthesizing,pan2025tokenhsi,wu2024human,xiao2024unified,deng2025human} use rule-based systems or LLMs/VLMs as planners with physics-based policies to guide agent completing long-horizon tasks.
While existing methods have achieved substantial progress, they predominantly address single-agent interactions with object and largely overlook multi-character scenarios. In contrast, our work focuses on a collaborative dual-humanoid setting involving coordinated object carrying and manipulation.

\begin{figure*}[h]
  \centering
  \includegraphics[width=\linewidth]{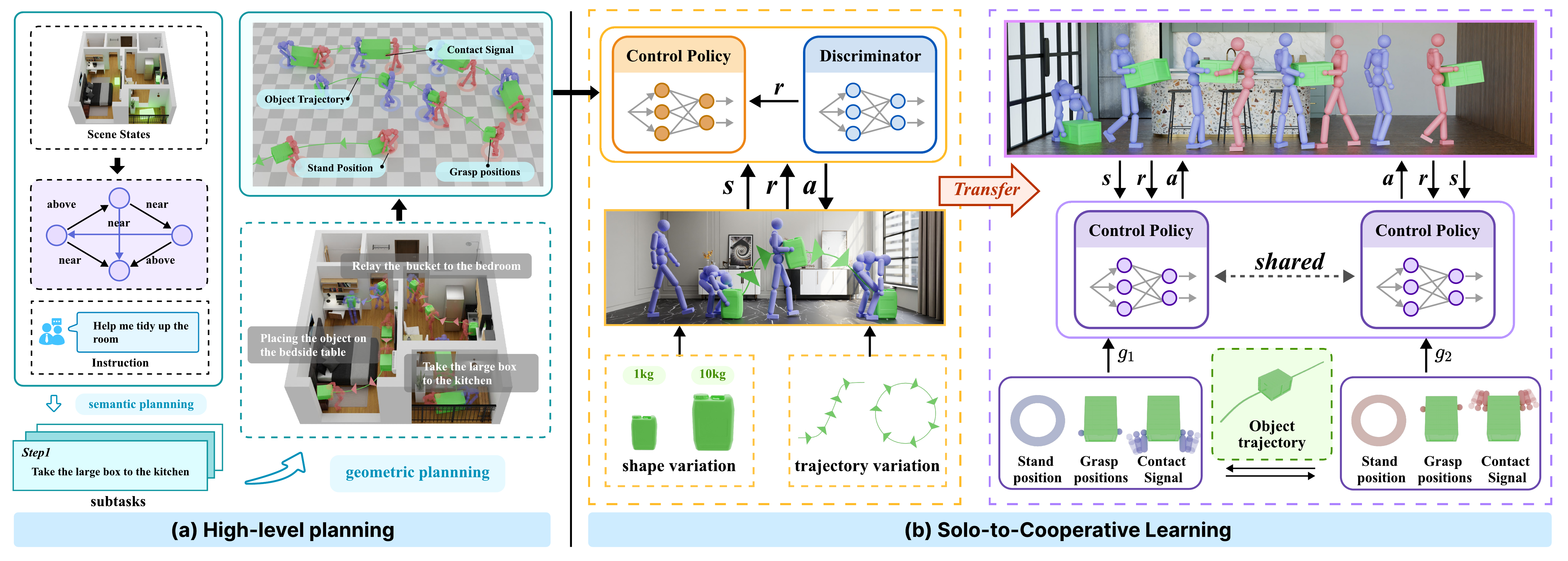}
  \vspace{-7mm}
  \caption{Overview. Given a long-horizon collaborative manipulation task, the strategy generation module generates high-level coordination cues, including object trajectories, standing positions, contact positions, and grasp signals. Conditioned on these cues, a shared policy controls all agents to execute coordinated object manipulation.}
  \vspace{-3mm}
  \label{fig:pipeline}
\end{figure*}
\section{Method}\label{sec:Method}

As shown in \cref{fig:pipeline}, this paper introduces a framework for multi-agent collaborative object manipulation across diverse objects and varied collaboration types in complex scenarios. Since enabling agents to understand how to collaborate is crucial for successful task completion, we decouple the task into a Commander--Executor formulation. The Commander provides high-level collaborative strategies that specify how agents should coordinate their behaviors, while the Executor translates these strategies into physically plausible multi-agent interactions with the object. \cref{sec:Coordination} describes how to formulate collaborative strategies by integrating task semantics with scene geometry. \cref{sec:Single} introduces the training of a single-agent interaction policy to acquire basic object interaction skills. Building on this policy, \cref{sec:Dual} further explains how to combine the single-agent policy with the collaborative strategy to achieve successful manipulation of diverse objects.

\subsection{Semantic- and Geometry-Aware Manipulation Strategy}\label{sec:Coordination}
For collaborative manipulation, an explicit manipulation strategy is essential for reliable task execution. This motivation comes from real-world observations: when multiple agents manipulate an object together, the Commander provides a shared cooperative strategy that informs each agent not only of the final goal but also of how to coordinate with others during execution. If agents act based only on their own local intentions, their behaviors can easily become inconsistent, causing conflicts in motion, unstable object transport, or even complete task failure. 

Based on this observation, we propose a method that jointly integrates semantic understanding with geometric reasoning. Specifically, we construct a Commander to reason over the relationships among agents, objects, and task goals, and to infer a global manipulation strategy for the overall task. This strategy provides structured guidance for the agents, allowing them to coordinate their motions according to both the semantic requirements of the task and the geometric constraints of the scene.

\noindent\textbf{Scene Graph Construction.} We represent the configurations of the scene using scene graphs $\mathcal{G}=(\mathcal{V}, \mathcal{E})$, where nodes denote object instances and edges encode their spatial relations. Specifically, the node set $\mathcal{V}=\{v_i\}_{i=1}^{N}$ denotes the objects in the scene. Each node encodes the geometric and identity information of an object, denoted as $v_i=(b_i, p_i, \theta_i, id_i)$, where $b_i$, $p_i$, $\theta_i$, and $id_i$ represent its bounding box, position, orientation, and object identifier, respectively. For edge construction, we follow \cite{jia2024sceneverse} and classify pairwise spatial relationships from object coordinates into ten categories: \textit{support, embed, inside/place-in, hanging, above/below, near, in-front-of/behind, left/right, between, and aligned}.

\noindent\textbf{Strategy Generation.} We input the scene graph and task instruction into an LLM~\cite{liu2024deepseek}, leveraging its strong reasoning ability to generate diverse collaborative manipulation strategies. Due to the hallucination problem of LLMs, it is difficult for them to directly generate dense object trajectories. To address this, we decompose the strategy generation task into two steps. We first leverage the LLM for semantic-level planning, decomposing the complete task into multiple subtasks. After that, we employ A$^\ast$~\cite{hart1968formal} and hybrid A$^\ast$~\cite{dolgov2008practical} for geometric-level planning, converting the subtasks into concrete and executable goals. Specifically, we require the LLM to deeply understand the scene layout and task information, reason about which objects need to be moved, their destinations, and the manipulation modes based on semantic and geometric cues, and then decompose the full task into multiple subtasks. Each subtask is represented as a text-based tuple $e=(o_e, m_e, p_e, r_e, \theta_e, h_e)$, where $o_e$ denotes the object to be moved, $m_e$ is the manipulation mode, $p_e$ and $r_e$ is a textual description of the destination and relay position, $\theta_e$ is the target orientation, and $h_e$ describes the grasping direction of each character. Since directly asking the LLM to output coordinates is difficult, we instead let it describe positions using other static objects in the scene and then sample the actual coordinates via simulated annealing~\cite{doi:10.1126/science.220.4598.671}.

While the LLM provides high-level semantic reasoning, our method further incorporates scene geometry to ensure the feasibility of the generated subtasks. When executing each subtask, we first sample the contact position $\hat{p}_t^{contact}$ and the target root position for each character along the grasping direction based on the object's actual position. We then use A$^\ast$ to plan a path from each character's current location to the target root position and simplify the resulting path with the Ramer-Douglas-Peucker (RDP)~\cite{douglas1973algorithms} algorithm. After that, we treat the simplified waypoints as the target standing positions $\hat{p}_t^{standing}$ for the agents, guiding them to navigate toward the object step by step. For obstacle avoidance, we discretize the scene into a navigable map, where cells occupied by objects or outside the accessible region are marked as obstacles. We further inflate obstacle regions according to the character body size to ensure sufficient clearance for the agents during movement. As for the object trajectory, we adopt Hybrid A$^\ast$ for computation. When characters carry an object, the combined entity formed by them can become long and narrow, so simple inflation would lose too much information. Therefore, we model the human-object system as a rectangle and compute the path on the non-inflated map with Hybrid A$^\ast$. We then interpolate the path with specified linear velocity and maximum angular velocity, assign a specific height, and obtain the object trajectory $\{o_t^{ref}\}_{t=0}^n$, where $o_t^{ref}=(\hat{p}_t^{obj}, \hat{\theta}_t^{obj})$ represents the object's position and rotation. During the execution of each subtask, we employ a Finite State Machine (FSM) to schedule the phases of approaching, grasping, handover, carrying and releasing, and encode a grasp flag $\hat{c}_t$ for each character in each phase to indicate whether they should hold the object. These elements together form the complete collaborative manipulation strategy, which guides subsequent character carrying tasks.

\subsection{Single-Agent Manipulation with Sparse Observations}\label{sec:Single}
Multi-agent collaborative object manipulation data are often limited and costly to obtain. Moreover, directly training a policy in the multi-agent setting, especially for asymmetric objects, can easily lead to poor local minima, where the agents fail to coordinate effectively and cannot jointly accomplish the manipulation task. Therefore, we first train a single-agent object manipulation policy to acquire basic object interaction capabilities, providing a more stable foundation for subsequent collaborative manipulation.

\noindent\textbf{Policy Training with Augmentable Data.} The primary goal of the single agent is to control a humanoid to interact with an object and transport it along a specific trajectory. Although several datasets provide human object interactions, such as OMOMO~\cite{li2023object}, the recorded interactions are either too simple with very short trajectories, or are limited to small objects and do not easily transfer to other object sizes. To avoid being constrained by the limited trajectory diversity in existing datasets, we adopt custom-designed trajectories for policy training. Specifically, we define three basic motion primitives: \textit{walking with turns}, \textit{circular paths}, and \textit{spinning in place}. By composing these primitives, our training data can cover a diverse set of practical manipulation trajectories. Since the number of base trajectories is limited and their speeds are fixed, we apply random temporal scaling during trajectory sampling to vary the effective speed of the reference object motion. We also randomly rotate the trajectories, which forces the character to learn to interact with the object while moving forward, backward, or sideways.

To improve generalization across different object types, we randomize the size and mass of the manipulated objects during training, encouraging the agent to explore diverse interaction strategies for objects with different physical properties. In particular, since existing single-agent object interaction datasets typically contain relatively small objects, we further enlarge the object size while reducing its mass, allowing the agent to successfully manipulate larger objects in the single-agent setting and better transfer to collaborative manipulation of large objects. Contact positions are also randomly assigned within a plausible range, so that the character learns to grasp stably from varying configurations. Through these domain randomizations, our policy learns to manipulate diverse objects in a variety of ways.

\noindent\textbf{States and Actions.} The policy takes $o_t=(s_t, g_t)$ as input, where $s_t$ is the character's proprioceptive state, consisting of its root pose, joint rotations and velocities, as well as the object's pose and velocity and $g_t=(\hat{p}_t^{standing}, \hat{p}_t^{contact}, \hat{c}_{t:t+l}, o_{t:t+l}^{ref})$ represents the interaction strategy. Here $\hat{p}_t^{standing}$, $\hat{p}_t^{contact}$ and $\hat{c}_{t:t+l}$ define where, how, and when the character should grasp the object, while $o_{t:t+l}^{ref}$ defines how the object should be manipulated. Compared to tracking full-body reference motions, sparse observations provide a more flexible supervision signal for policy learning, allowing the agent to learn task-relevant behaviors without relying on detailed human interaction motion data. In addition, the trajectories provide temporal guidance on how the object should be carried, specifying not only the target location but also the intermediate motion process. This imposes stronger constraints on policy learning and reduces ambiguity in manipulation, which is important for enabling successful multi-agent collaboration in later tasks. At each time step, the policy network predicts target joint angles $a_t$, which are then converted into joint torques by PD controllers to drive the character in simulation.

\noindent\textbf{Reward Design.} Single-agent manipulation can be decomposed into three phases: approaching, grasping, and carrying the object. We design phase-specific rewards to guide the character in learning the complete task in a structured manner.

\begin{figure*}[h!]
  \centering
  \includegraphics[width=\linewidth]{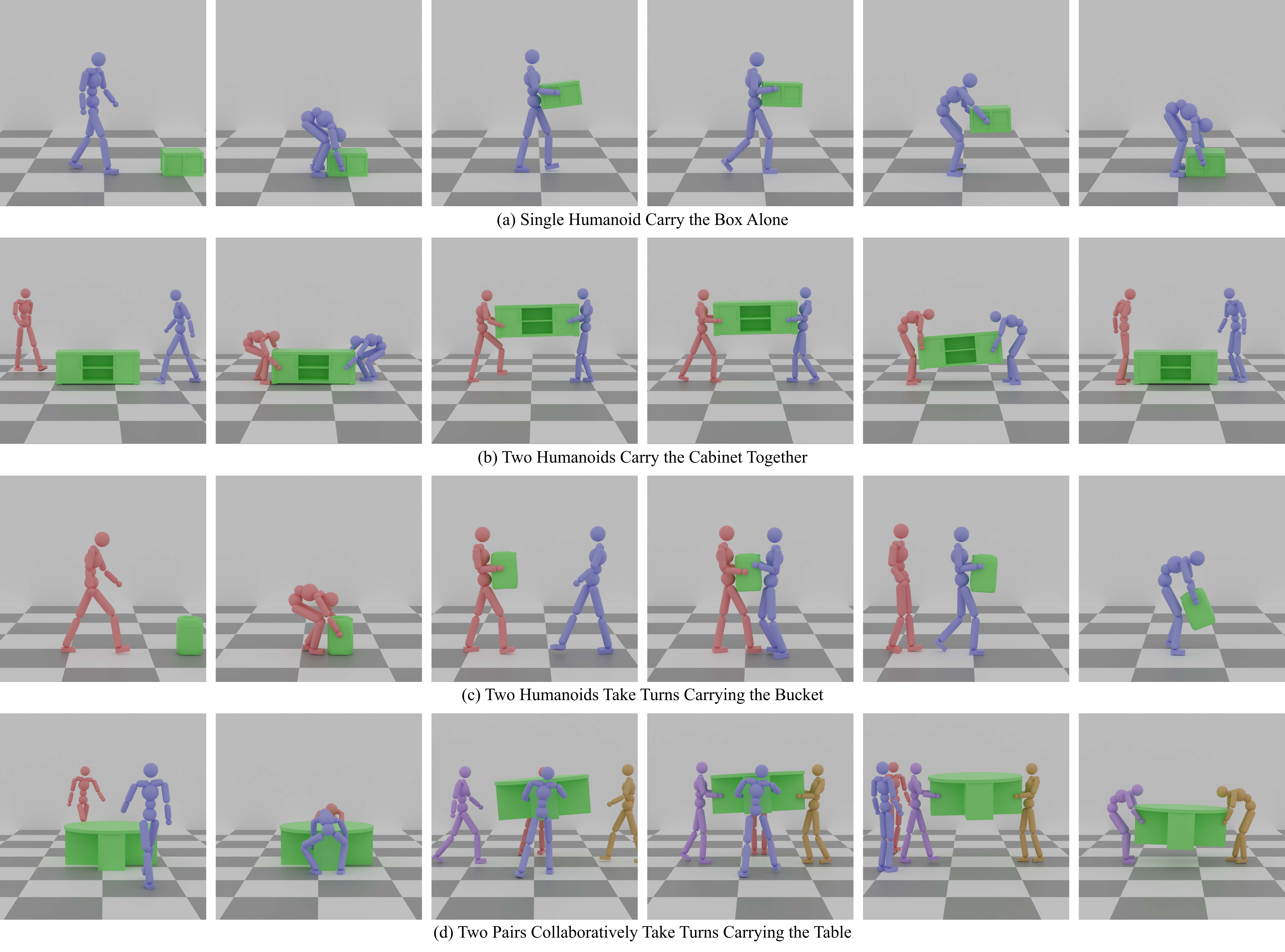}
  \vspace{-8mm}
  \caption{Our method can accomplish various carrying tasks, including single-humanoid carrying, two-humanoid cooperative carrying and relay carrying. It can also be extended to more roles to enable multiple humanoids cooperative relay carrying in groups.}
  \label{fig:task}
  \vspace{-2mm}
\end{figure*}

In addition to the AMP reward~\cite{peng2021amp}, we guide each character toward an appropriate stance using the designated standing position specified by the manipulation strategy, from which the character can directly reach and grasp the object. We therefore define a walking reward $r_{\mathrm{walk}}$ based on the distance, orientation, and speed of character's root relative to the target standing position. Specifically, let $u_t^{root}$ denote the root orientation, $\hat{u}_t^{standing}$ be the direction from the character to the target standing point, $v_t^{root}$ be the root velocity and $v^*$ be the desired target speed. We then formulate the walking reward as:
\begin{equation}
\begin{split}
r_{walk}= & \exp(\|p_t^{root}-\hat{p}_t^{standing}\|^2) + \\
& \|u_t^{root} \cdot \hat{u}_t^{standing}\|^2 + \\
& \exp(\|v^*-u_t^{root} \cdot v_t^{root}\|^2)
\end{split}
\label{eq:walk}
\end{equation}
Because the transported objects vary in shape, we specify distinct left and right hand contact positions for each character. These targets must be not only kinematically plausible but also symmetric in height to allow stable carrying. The grasp reward $r_{grasp}$ is defined by the distances between the hands and their target contact locations if the agent is expected the contact the object:
\begin{equation}
r_{grasp}=mean({exp(\|p_t^{hands}-\hat{p}_t^{contact}\|^2)}).
\label{eq:grasp}
\end{equation}
Finally, the character must carry the object along the given trajectory. We compute the reference position and rotation of the object at each time step and measure the deviation of the current object pose from the reference. The carrying reward $r_{carry}$ is defined as:
\begin{equation}
r_{carry}=exp(\|p_t^{obj}-\hat{p}_t^{obj}\|^2) \times exp(|{\theta}_t^{obj}-\hat{\theta}_t^{obj}|).
\label{eq:carry}
\end{equation}

Although phase decomposition reduces the difficulty of policy learning, phase transitions become challenging, since the policy must satisfy the activation condition while maintaining stable task progress. For example, an agent may remain near the object without initiating object-trajectory tracking. In this case, the episode does not terminate, allowing the agent to accumulate walking rewards over time and eventually converge to an undesirable local optimum. To address this obstacle, we introduce an activation reward:
\begin{equation}
r_{\mathrm{activate}} =
\begin{cases}
1 & \text{if the trajectory-following task is initiated}, \\
0 & \text{otherwise}.
\end{cases}
\end{equation}
Once the distance between the character's root and the target standing position falls below a predefined threshold, the trajectory-foll-owing task is initiated and the activation reward $r_{activate}$ is enabled. The details of reward functions can be found in the supplementray materials.

\noindent\textbf{Early Termination.} During training we observed that $r_{activate}$ alone is still insufficient to escape the local optimum. We therefore monitor the character's progress toward the target over a 1-second window. If the progress is less than 0.2\textit{m}, we terminate the episode early to encourage the character to approach the object. During the carrying phase, we also apply early termination when the object deviates excessively from the reference trajectory, avoiding unnecessary computation on unpromising rollouts. To accelerate learning in the carrying phase, we initialize the character near the designated standing position with probability 0.8, and from a random position otherwise, allowing the policy to experience grasping and carrying states more frequently.

\begin{figure*}[h!]
  \centering
  \includegraphics[width=\linewidth, alt={Qualitative comparisons with physics-based methods.}]{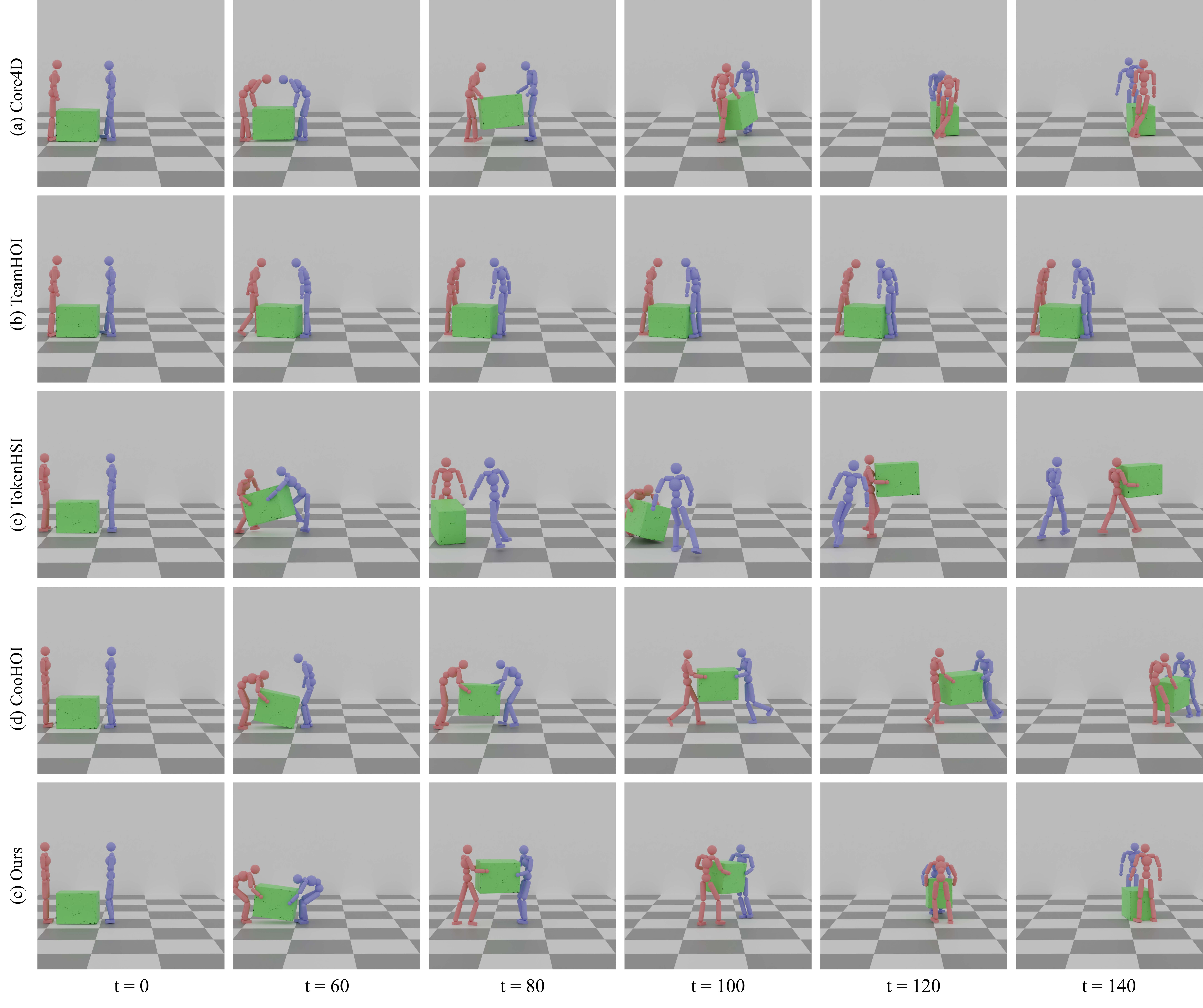}
  \caption{Qualitative comparisons with physics-based methods. The motion in Core4D dataset contains noticeable artifacts in reconstructed motions and object interactions. TokenHSI struggles to adapt from single-agent interaction to multi-agent collaboration, while TeamHOI often fails to grasp objects due to its fixed contact patterns. CooHOI can produce cooperative behaviors but supports only limited manipulation patterns. In contrast, our method generates more natural and coordinated collaborative manipulation, with adaptive grasping behaviors and smoother humanoid-object interactions.}
  \label{fig:comparison}
\end{figure*}

\subsection{Solo-to-Cooperative Learning}\label{sec:Dual}
Compared with single-agent object manipulation, multi-agent object manipulation requires each agent not only to interact with the object, but also to coordinate its actions in response to the dynamic behaviors of other agents. Such interactions can take various forms, with cooperative carrying and handover being two common scenarios. These collaborative settings introduce several additional challenges:
\begin{itemize}
\item agents may collide with each other in the absence of proper planning or mutual awareness;
\item agents may fail to synchronize their actions without unified coordination;
\item smooth handover requires clear transfer intent and interaction cues.
\end{itemize}

To this end, our designs in strategy generation~(\cref{sec:Coordination}) and single-agent manipulation~(\cref{sec:Single}) can be readily transferred to the multi-agent setting to address these challenges. To avoid inter-agent conflicts, we assign each agent a target standing position $\hat{p}_t^{standing}$ and contact position $\hat{p}t^{contact}$, encouraging the agents to approach and grasp distinct regions of the object. For coordination, both agents follow a shared object reference trajectory $o_{t:t+l}^{ref}$, which provides a unified target for synchronizing their manipulation behaviors. For object transfer, the standing positions $\hat{p}_t^{standing}$ and grasp signals $\hat{c}_{t:t+l}$ specify where each agent should move and when it should grasp or release the object, enabling a smooth handover.

With these designs, we can directly fine-tune the trained single-agent policy on multi-agent collaborative tasks without modifying the policy architecture. All agents are controlled by the same shared policy, while receiving agent-specific observations that reflect their assigned roles and local states. Because the policy has already learned basic object manipulation skills, multi-agent fine-tuning becomes less susceptible to undesirable local optima and achieves improved training efficiency. 

Compared with the single-agent setting, we train the joint carrying task with larger and heavier objects while retaining random object augmentation for generalization. The activation condition is also modified such that object trajectory following is activated only when all agents are sufficiently close to the object. For the relay task, we designate one agent as the giver and the other as the receiver. The giver approaches, grasps, and carries the object toward a handover location planned by the LLM. Meanwhile, the receiver moves to the planned handover location and waits for the transfer. Once the handover is completed, the receiver takes over the object and carries it to the final target. By assigning different standing positions and grasp signals to the giver and receiver, we can specify the handover process without introducing additional task-specific modules. In the relay task, we use the same object size and mass ranges as in the single-agent setting.

By using the object trajectory as the only communication signal among agents, our solo-to-cooperative learning enables an arbitrary number of agents to collaborate on the same manipulation task. Since the underlying task objectives are consistent between single-agent and multi-agent scenarios, we directly fine-tune the policy pretrained on the single-agent task, using exactly the same reward functions and environment settings.
\section{Experiments}\label{sec:Experiments}

\begin{table*}[htbp]
\centering
\caption{Comparisons with physics-based methods. Our method achieves a higher success rate while maintaining lower tracking errors, demonstrating more reliable and accurate cooperative manipulation.}
\vspace{-2mm}
\label{tab:physics}
\resizebox{\textwidth}{!}{
\begin{tabular}{l c c c c c c c c c c c c}
\toprule
\multirow{2}{*}{Method} & \multicolumn{6}{c}{Core4D-Trajectory} & \multicolumn{6}{c}{Bezier-Trajectory} \\
\cmidrule(lr){2-7} \cmidrule(lr){8-13}
& SR.$\uparrow$ & DE.$\downarrow$ & OE.$\downarrow$ & DTE.$\downarrow$ & OTE.$\downarrow$ & Contact Acc.$\uparrow$ 
& SR.$\uparrow$ & DE.$\downarrow$ & OE.$\downarrow$ & DTE.$\downarrow$ & OTE.$\downarrow$ & Contact Acc.$\uparrow$  \\
\midrule
TokenHSI~\cite{pan2025tokenhsi} & 20.63 & 0.279 & 0.496 & 0.454 & 0.860 & 56.82 & 5.990 & 0.126 & 0.512 & 0.247 & 0.343 & 18.33 \\
CooHOI~\cite{gao2024coohoi} & 39.47 & \textbf{0.098} & 0.220 & 0.253 & 0.211 & 78.38 & 35.42 & 0.102 & 0.192 & 1.135 & 0.471 & 44.22 \\
TeamHOI~\cite{lionar2026teamhoi} & 7.900 & 0.290 & 0.267 & 0.303 & 0.339 & 27.90 & 19.33 & 0.096 & 0.455 & 0.190 & 0.411 & 6.810 \\
\textbf{Ours} & \textbf{80.26} & 0.157 & \textbf{0.147} & \textbf{0.122} & \textbf{0.133} & \textbf{86.23} & \textbf{86.50} & \textbf{0.083} & \textbf{0.162} & \textbf{0.088} & \textbf{0.149} & \textbf{98.17} \\
\bottomrule
\end{tabular}
}
\vspace{-3mm}
\end{table*}

In this section, we quantitatively evaluate the capability of the low-level policy when two agents collaboratively transport an object, as well as the overall algorithm's performance in terms of planning and transportation within complex scenarios. Since no baseline method supports handover, a fair quantitative comparison is not feasible, and we therefore present qualitative results in the supplementary video for handover task. We employ PPO~\cite{schulman2017proximal} as the reinforcement learning algorithm for low-level policy training and DeepSeek-V4-Pro~\cite{{deepseekai2026deepseekv4highlyefficientmilliontoken}} as the high-level planner. Detailed training configurations are provided in the supplementary materials.

\subsection{Datasets and Metrics}\label{sec:datasets}
\noindent\textbf{Datasets.} Due to the scarcity of large-scale multi-person collaborative object interaction data, our training minimizes reliance on such data. We use single-person motion-only datasets such as AMASS \cite{mahmood2019amass} to train the adversarial motion prior, providing diverse human motions for better generalization. For evaluation, we use Core4D~\cite{liu2025core4d}, extracting object meshes and trajectories while filtering out those with abnormal velocities or accelerations to construct a reliable benchmark. Since the remaining trajectories are relatively simple, we additionally generated 200 Bezier curves, with 50 curves for each setting of 2 to 5 control points sampled within a 10$\times$10 m area. For each generated trajectory, the speed was randomly sampled from 0.8 to 1.3 m/s, and the trajectory was randomly rotated to allow the object to move in various directions. For experiments in complex scenarios, we construct two spatial layouts: a simple two-bedroom one-living-room apartment (2B1L) and a more complex L-shaped configuration. These layouts are designed to evaluate the algorithm's geometric planning and policy control capabilities in complex environments. Each layout is divided into four regions: one starting zone and three target zones. Three objects are placed in the starting zone, each corresponding to one of the target zones. For each layout, we design one precise instruction and one vague instruction to assess the algorithm's semantic understanding ability.

\noindent\textbf{Metrics.} For low-level carrying tasks, we report Success Rate (SR.), Distance Error (DE), Orientation Error (OE), Distance Tracking Error (DTE), and Orientation Tracking Error (OTE). A trial is considered successful if the final $xy$-plane distance error is below 0.3 m and the final orientation error is below $30^\circ$. DE and OE measure the final $xy$-plane distance error in meters and orientation error in radians, respectively, while DTE and OTE measure their average per-frame tracking errors along the reference trajectory. For human-object interaction, following \cite{xu2026stability}, we compute Contact Accuracy (Contact Acc.) to evaluate whether hand-object contacts are correctly generated. For the experiments in complex scenarios, we evaluate Semantic Success Rate (SSR.) and Carrying Success Rate (CSR.) over the three objects. A trial is considered semantically successful if the LLM correctly plans the placement of each object into its designated target zone and sample a reachable position within that zone.  Carrying success is defined as successfully moving an object to the location designed by the high-level plan.

\subsection{Comparison to State-of-the-Art Methods}\label{sec:Comparison}

To comprehensively evaluate our method on physical control, we compare it with state-of-the-art physics-based methods. We consider TeamHOI~\cite{lionar2026teamhoi}, CooHOI~\cite{gao2024coohoi}, and TokenHSI~\cite{pan2025tokenhsi} as representative baselines for multi-agent and human-object interaction. TeamHOI and CooHOI are recent physics-based frameworks for cooperative motion generation. Since Token-HSI is originally designed for single-agent human-object interaction, we adapt it to our setting by allowing two agents to interact with the same object.

\cref{tab:physics} presents the comparison between our method and existing physics-based methods. We evaluate on a variety of objects from Core4D dataset with different shapes and sizes to demonstrate the adaptability, while also randomizing object masses in the range of 10 kg to 25 kg. We use the original trajectories from Core4D and additionally generate object trajectories using Bezier curves to further evaluate the generalization ability of our method. The results show that our method outperforms prior works in both Success Rate and Tracking Error. As illustrate in \cref{fig:comparison}, TokenHSI~\cite{pan2025tokenhsi}, originally designed for single-agent interaction, struggles to coordinate two agents when adapted to human-human-object interaction. CooHOI~\cite{gao2024coohoi} supports dual agents but relies on sparse position as target, which makes precise trajectory tracking difficult to achieve. TeamHOI~\cite{lionar2026teamhoi} generates cooperative behaviors, but its fixed grasping strategy limits generalization across objects. In contrast, our method assigns role-specific grasping locations through the planned coordination strategy, enabling robust adaptation to diverse object shapes and sizes.

In addition, \cref{fig:task} demonstrates the versatility of our method across different collaborative tasks. Our method can control a single agent to manipulate an object, coordinate two agents for joint carrying and relay-style handover, and further scale to multi-agent collaboration by composing multiple instances of the same shared policy. This compositional design allows more complex cooperative behaviors, such as group-based relay manipulation, without introducing task-specific policy architectures.

\subsection{Long Horizon Tasks}\label{Long Horizon Tasks}
We evaluate our algorithm's planning and execution performance in two scenarios. In each scenario, we initialize the positions of three objects within the starting zone and use both precise and vague instructions. The results show that our method achieve a 90\% success rate in understanding semantic information and planning across all scenarios, which is attribute to our decoupling of semantic and geometric reasoning. This design significantly alleviated the burden on the LLM. In the subsequent execution phase, the success rates for sequentially carrying one, two, and three objects are 0.92, 0.82, and 0.67, respectively. Although our algorithm does not perceive the surrounding environment, it effectively avoids collisions between the agents and the scene through geometric planning and precise control centered on object trajectories. Given a single-carrying success rate of 0.9, the expected success rates for two and three consecutive transports would be 0.81 and 0.73. Our results are close to these values, indicating that task switching has only a minor impact on overall performance.

\begin{table}[t]
\centering
\caption{Quantitative results for long-horizon task in complete scenarios. CSR\#k denotes the success rate of continuously carrying k objects.}
\label{tab:Ablation}
\begin{tabular}{lcccc}
\toprule
Scene & SSR$\uparrow$ & CSR\#1$\uparrow$ & CSR\#2$\uparrow$ & CSR\#3$\uparrow$ \\
\midrule
2B1L & 97.22 & 94.44 & 86.11 & 69.44 \\
L-shaped & 94.44 & 88.89 & 77.78 & 63.89 \\
\midrule
Average & 95.83 & 91.67 & 81.95 & 66.67 \\
\bottomrule
\end{tabular}
\vspace{-4mm}
\end{table}

\begin{figure}[h]
  \centering
  \includegraphics[width=\linewidth, alt={Task Reward Curves.}]{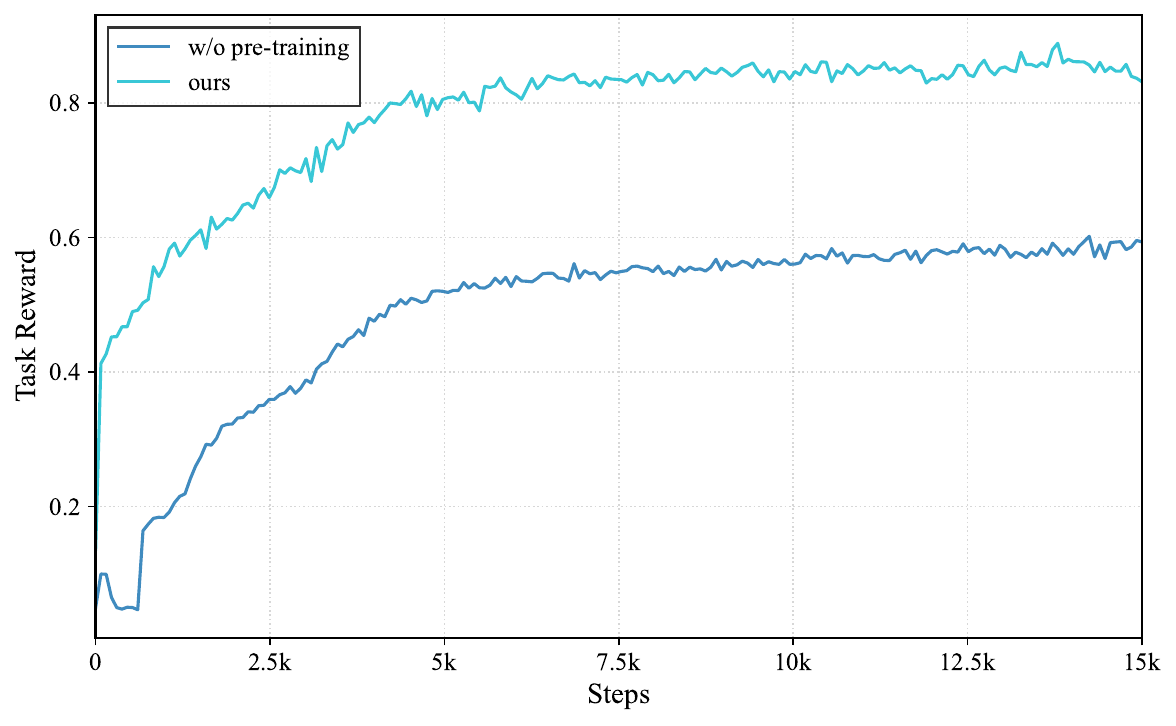}
  \vspace{-8mm}
  \caption{Training the policy directly on the multi-agent collaborative task without prior single-agent pretraining tends to cause early overfitting, as reflected by degraded reward stability and limited generalization.}
  \vspace{-5mm}
  \label{fig:solo-to-coop}
\end{figure}

\begin{figure*}[h!]
  \centering
  \includegraphics[width=\linewidth]{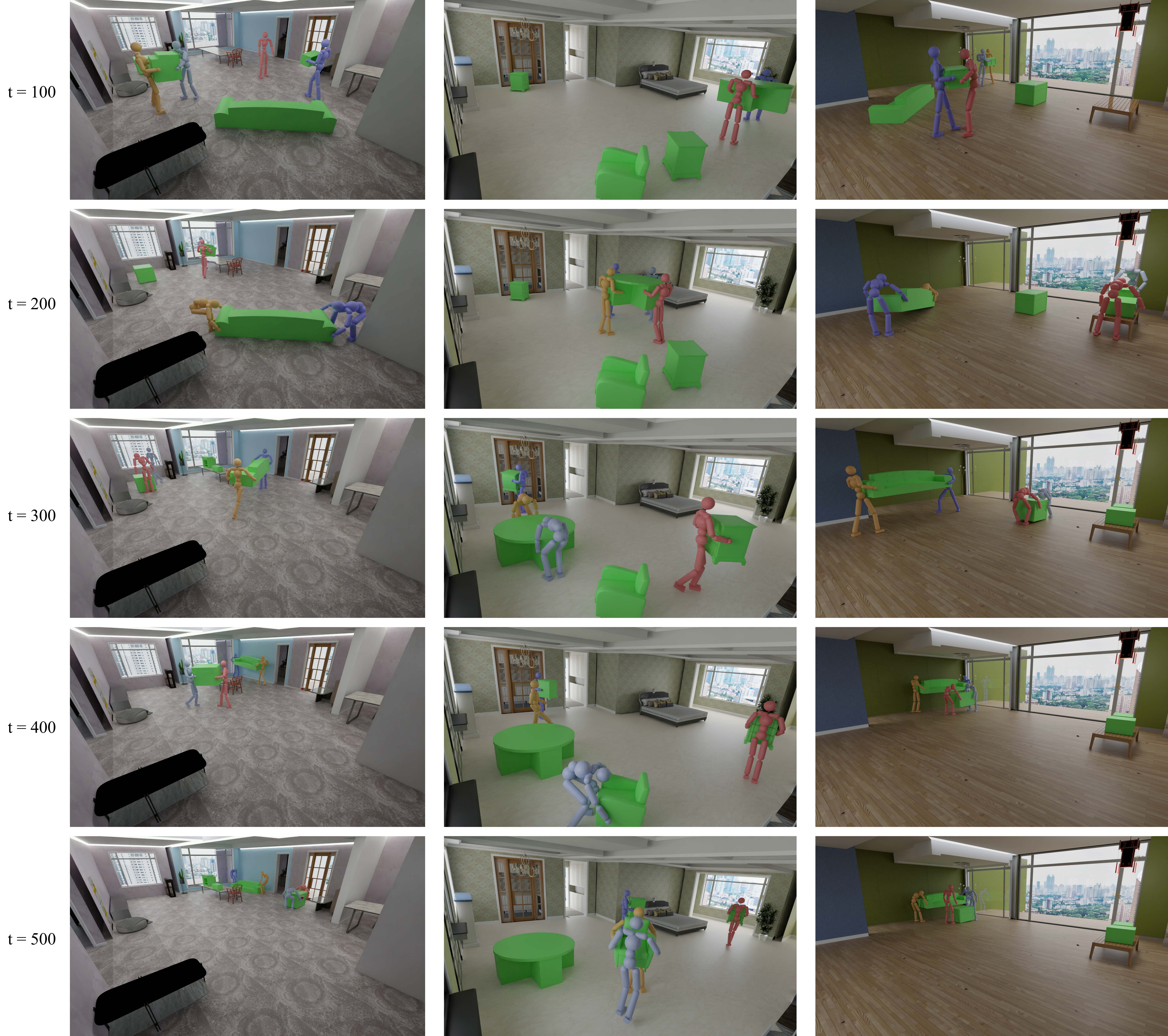}
  \caption{Our method enables diverse object manipulation in complex scenes by combining task- and geometry-aware strategy planning with physics-based policy execution. Given different scene layouts and object goals, it can generate feasible coordination strategies and control multiple humanoids to perform long-horizon interactions, including cooperative carrying, relay-style handover, and group-based manipulation across objects with different shapes and sizes.}
  \label{fig:moive}
\end{figure*}

\subsection{Solo-to-Cooperative Learning}\label{sec:Solo-to-Cooperative}
We ablate the training strategy of the manipulation policy. As shown in \cref{fig:solo-to-coop}, directly training the policy on the multi-agent cooperative task without single-agent pretraining leads to severe overfitting. In this case, one agent may quickly learn to complete its assigned motion, while the other consistently fails, preventing the agents from accomplishing the cooperative task together. Conversely, when the policy is trained only on single-agent object interaction and is not fine-tuned on the multi-agent cooperative task, it fails to acquire cooperative behaviors. Since the policy has not learned to adjust its own state according to the object state and the partner's behavior, it cannot coordinate with the other agent during object manipulation. The quantitative results in \cref{tab:Ablation} further confirm this observation: removing either single-agent pretraining or multi-agent fine-tuning causes the success rate to drop to zero, demonstrating that both stages are essential for robust cooperative manipulation.

\begin{table}[t]
\centering
\caption{Quantitative results for ablation studies. Removing key components significantly degrades performance, demonstrating their importance for stable and successful collaborative manipulation.}
\vspace{-1mm}
\label{tab:Ablation}
\resizebox{\columnwidth}{!}{
\begin{tabular}{lccccc}
\toprule
Method & SR.$\uparrow$ & DE.$\downarrow$ & OE.$\downarrow$ & DTE.$\downarrow$ & OTE.$\downarrow$ \\
\midrule
w/o single-agent pretraining & 0 & - & - & - & - \\
w/o multi-agent fine-tuning & 0 & - & - & - & - \\
w/o object-trajectory & 0.170 & 0.192 & 0.293 & 1.047 & 0.712 \\
w/o root pos & 76.08 & 0.118 & 0.159 & 0.102 & 0.137 \\
w/o contact pos & 79.67 & 0.140 & 0.074 & 0.105 & 0.123 \\
w/o contact signal & 78.91 & 0.143 & 0.146 & 0.096 & 0.146 \\
w/o shape variation & 23.83 & 0.087 & 0.054 & 0.093 & 0.140 \\
Ours & 86.50 & 0.083 & 0.162 & 0.088 & 0.149 \\
\bottomrule
\end{tabular}
}
\vspace{-4mm}
\end{table}

\subsection{Collaborative Strategy from Commander}\label{sec:Strategy}
For cooperative manipulation, it is crucial for all agents to understand not only the final task objective but also their roles in achieving it. Therefore, a commander is needed to provide an explicit collaborative strategy. In our framework, we use object trajectories as this strategy, which guides the agents toward coordinated object transport. In \cref{tab:Ablation}, we replace the trajectory guidance with only the target position. Under this setting, each agent only knows where the object should be transported, but does not receive information on how to cooperate with the other agent during manipulation. The results show that the task success rate drops substantially, demonstrating that explicit trajectory-based strategy from the commander is essential for effective multi-agent collaboration.

\subsection{Synchronized Grasping via Contact Signal}\label{sec:Signal}
Whether the agent can maintain stable contact with the object is crucial for carrying tasks. In this framework, we explicitly synchronize the timing of contact and release between two agents via contact signals. In \cref{tab:Ablation}, we remove the explicit synchronization provided by the contact signals. Under this setting, the success rate drops significantly while other metrics show no major differences, indicating that contact signals effectively help agents achieve stable grasping.

\subsection{Generalization through Shape Variation}\label{sec:Variation}
Generalization to diverse objects is essential in practice. To improve the policy's generalization capability, we randomize the shape of objects during training, thereby enhancing robustness across different objects. In \cref{tab:Ablation}, we ablate the generalization to object shape. Under this setting, the policy struggles to interact with unseen objects, resulting in a sharp drop in success rate. This indicates that generalizing over object shapes during training is crucial for improving robustness.
\section{Discussion and Limitations}\label{sec:Conclusion}
In this paper, we present a scalable Commander-Executor framework for long-horizon collaborative object manipulation. The Commander jointly reasons about task semantics and scene geometry to generate feasible coordination strategies, while the Executor learns a physics-based policy conditioned on strategy cues to produce coordinated humanoid-object interactions. To reduce reliance on scarce multi-agent interaction data, we introduce a solo-to-coop-erative learning scheme that trains a single-agent policy with augmented objects, contacts, and trajectories, and transfers it to multi-agent collaboration through a shared policy. Experiments show that our method supports diverse collaboration types, generalizes across object categories, and coordinates multiple agents in complex long-horizon tasks.

Although the proposed method enables scalable collaborative object manipulation, it still has several limitations. For the low-level policy, the use of simplified spherical hand representations restricts dexterous manipulation, making it difficult to handle slender objects such as baseball bats or ladders. It also falls short when it comes to executing highly skill-demanding maneuvers. In addition, during collaborative carrying task, if the object's mass and size are sufficiently small, it may end up being controlled by just one agent. This happens because our policy does not distinguish between single-agent and dual-agent carrying in its input. However, as the mass and size gradually increase, this issue completely disappears. For the high-level commander, while our Commander considers scene geometry for strategy planning, the current physics-based execution mainly focuses on humanoid-object interaction and does not fully model rich physical interactions with complex environments, such as stairs, slopes, or deformable obstacles. Future work will explore more detailed hand-object contact models and tighter integration between scene-aware planning and environment-aware control.

\section*{Acknowledgments}
This work was supported in part by the National Natural Science Foundation of China (No. 62631020 and No. 62601569), and the Open Project Program of State Key Laboratory of Virtual Reality Technology and Systems, Beihang University (No.VRLAB2026C04).

\bibliographystyle{IEEEtran}
\bibliography{main}

@inproceedings{mahmood2019amass,
  title={AMASS: Archive of motion capture as surface shapes},
  author={Mahmood, Naureen and Ghorbani, Nima and Troje, Nikolaus F and Pons-Moll, Gerard and Black, Michael J},
  booktitle={Proceedings of the IEEE/CVF international conference on computer vision},
  pages={5442--5451},
  year={2019}
}

@inproceedings{lionar2026teamhoi,
  title={TeamHOI: Learning a Unified Policy for Cooperative Human-Object Interactions with Any Team Size},
  author={Lionar, Stefan and Lee, Gim Hee},
  booktitle={Proceedings of the IEEE/CVF Conference on Computer Vision and Pattern Recognition (CVPR)},
  year={2026}
}

@inproceedings{xu2026stability,
  title={Stability-Driven Motion Generation for Object-Guided Human-Human Co-Manipulation},
  author={Xu, Jiahao and Yuan, Xiaohan and Wu, Xingchen and Xu, Chongyang and Li, Kun and Huang, Buzhen},
  booktitle={Proceedings of the IEEE/CVF Conference on Computer Vision and Pattern Recognition (CVPR)},
  year={2026}
}

@inproceedings{liu2025core4d,
title={Core4d: A 4d human-object-human interaction dataset for collaborative object rearrangement},
author={Liu, Yun and Zhang, Chengwen and Xing, Ruofan and Tang, Bingda and Yang, Bowen and Yi, Li},
booktitle={Proceedings of the Computer Vision and Pattern Recognition Conference},
pages={1769--1782},
year={2025}
}

@inproceedings{daiya2025collage,
title={COLLAGE: Collaborative human-agent interaction generation using hierarchical latent diffusion and language models},
author={Daiya, Divyanshu and Conover, Damon and Bera, Aniket},
booktitle={2025 IEEE International Conference on Robotics and Automation (ICRA)},
pages={8203--8210},
year={2025},
organization={IEEE}
}

@inproceedings{ji2025onlinehoi,
title={OnlineHOI: Towards Online Human-Object Interaction Generation and Perception},
author={Ji, Yihong and Liu, Yunze and Zhuo, Yiyao and Yu, Weijiang and Ma, Fei and Huang, Joshua Zhexue and Yu, Fei},
booktitle={Proceedings of the 33rd ACM International Conference on Multimedia},
pages={9395--9403},
year={2025}
}

@inproceedings{he2025syncdiff,
title={Syncdiff: Synchronized motion diffusion for multi-body human-object interaction synthesis},
author={He, Wenkun and Liu, Yun and Liu, Ruitao and Yi, Li},
booktitle={Proceedings of the IEEE/CVF International Conference on Computer Vision},
pages={11731--11743},
year={2025}
}

@article{vaillant2016multi,
  title={Multi-character physical and behavioral interactions controller},
  author={Vaillant, Joris and Bouyarmane, Karim and Kheddar, Abderrahmane},
  journal={IEEE transactions on visualization and computer graphics},
  volume={23},
  number={6},
  pages={1650--1662},
  year={2016},
  publisher={IEEE}
}

@article{liu2024deepseek,
  title={Deepseek-v3 technical report},
  author={Liu, Aixin and Feng, Bei and Xue, Bing and Wang, Bingxuan and Wu, Bochao and Lu, Chengda and Zhao, Chenggang and Deng, Chengqi and Zhang, Chenyu and Ruan, Chong and others},
  journal={arXiv preprint arXiv:2412.19437},
  year={2024}
}

@inproceedings{zhang2023simulation,
title={Simulation and retargeting of complex multi-character interactions},
author={Zhang, Yunbo and Gopinath, Deepak and Ye, Yuting and Hodgins, Jessica and Turk, Greg and Won, Jungdam},
booktitle={ACM SIGGRAPH 2023 Conference Proceedings},
pages={1--11},
year={2023}
}

@article{gao2024coohoi,
title={Coohoi: Learning cooperative human-object interaction with manipulated object dynamics},
author={Gao, Jiawei and Wang, Ziqin and Xiao, Zeqi and Wang, Jingbo and Wang, Tai and Cao, Jinkun and Hu, Xiaolin and Liu, Si and Dai, Jifeng and Pang, Jiangmiao},
journal={Advances in Neural Information Processing Systems},
volume={37},
pages={79741--79763},
year={2024}
}

@inproceedings{xu2025perceiving,
  title={Perceiving and Acting in First-Person: A Dataset and Benchmark for Egocentric Human-Object-Human Interactions},
  author={Xu, Liang and Yang, Chengqun and Lin, Zili and Xu, Fei and Liu, Yifan and Xu, Congsheng and Zhang, Yiyi and Qin, Jie and Sheng, Xingdong and Liu, Yunhui and others},
  booktitle={Proceedings of the IEEE/CVF International Conference on Computer Vision},
  pages={12535--12548},
  year={2025}
}

@article{na2025learning,
  title={Learning to Generate Human-Human-Object Interactions from Textual Descriptions},
  author={Na, Jeonghyeon and Baik, Sangwon and Lee, Inhee and Lee, Junyoung and Joo, Hanbyul},
  journal={arXiv preprint arXiv:2511.20446},
  year={2025}
}

@article{schulman2017proximal,
  title={Proximal policy optimization algorithms},
  author={Schulman, John and Wolski, Filip and Dhariwal, Prafulla and Radford, Alec and Klimov, Oleg},
  journal={arXiv preprint arXiv:1707.06347},
  year={2017}
}

@article{peng2018deepmimic,
  title={Deepmimic: Example-guided deep reinforcement learning of physics-based character skills},
  author={Peng, Xue Bin and Abbeel, Pieter and Levine, Sergey and Van de Panne, Michiel},
  journal={ACM Transactions On Graphics (TOG)},
  volume={37},
  number={4},
  pages={1--14},
  year={2018},
  publisher={ACM New York, NY, USA}
}

@inproceedings{zhang2025physics,
  title={Physics-Based Motion Imitation with Adversarial Differential Discriminators},
  author={Zhang, Ziyu and Bashkirov, Sergey and Yang, Dun and Shi, Yi and Taylor, Michael and Peng, Xue Bin},
  booktitle={Proceedings of the SIGGRAPH Asia 2025 Conference Papers},
  pages={1--12},
  year={2025}
}

@article{mu2025smp,
  title={SMP: Reusable Score-Matching Motion Priors for Physics-Based Character Control},
  author={Mu, Yuxuan and Zhang, Ziyu and Shi, Yi and Matsumoto, Minami and Imamura, Kotaro and Tevet, Guy and Guo, Chuan and Taylor, Michael and Shu, Chang and Xi, Pengcheng and others},
  journal={arXiv preprint arXiv:2512.03028},
  year={2025}
}

@article{wang2023physhoi,
  author    = {Wang, Yinhuai and Lin, Jing and Zeng, Ailing and Luo, Zhengyi and Zhang, Jian and Zhang, Lei},
  title     = {PhysHOI: Physics-Based Imitation of Dynamic Human-Object Interaction},
  journal   = {arXiv preprint arXiv:2312.04393},
  year      = {2023},
}

@inproceedings{Luo2023PerpetualHC,
    author={Zhengyi Luo and Jinkun Cao and Alexander W. Winkler and Kris Kitani and Weipeng Xu},
    title={Perpetual Humanoid Control for Real-time Simulated Avatars},
    booktitle={International Conference on Computer Vision (ICCV)},
    year={2023}
}

@InProceedings{Wang_2025_CVPR,
    author    = {Wang, Yinhuai and Zhao, Qihan and Yu, Runyi and Tsui, Hok Wai and Zeng, Ailing and Lin, Jing and Luo, Zhengyi and Yu, Jiwen and Li, Xiu and Chen, Qifeng and Zhang, Jian and Zhang, Lei and Tan, Ping},
    title     = {SkillMimic: Learning Basketball Interaction Skills from Demonstrations},
    booktitle = {Proceedings of the Computer Vision and Pattern Recognition Conference (CVPR)},
    month     = {June},
    year      = {2025},
    pages     = {17540-17549}
}

@inproceedings{xu2025intermimic,
  title={Intermimic: Towards universal whole-body control for physics-based human-object interactions},
  author={Xu, Sirui and Ling, Hung Yu and Wang, Yu-Xiong and Gui, Liang-Yan},
  booktitle={Proceedings of the Computer Vision and Pattern Recognition Conference},
  pages={12266--12277},
  year={2025}
}

@article{peng2021amp,
  title={Amp: Adversarial motion priors for stylized physics-based character control},
  author={Peng, Xue Bin and Ma, Ze and Abbeel, Pieter and Levine, Sergey and Kanazawa, Angjoo},
  journal={ACM Transactions on Graphics (ToG)},
  volume={40},
  number={4},
  pages={1--20},
  year={2021},
  publisher={ACM New York, NY, USA}
}

@inproceedings{pan2025tokenhsi,
  title={Tokenhsi: Unified synthesis of physical human-scene interactions through task tokenization},
  author={Pan, Liang and Yang, Zeshi and Dou, Zhiyang and Wang, Wenjia and Huang, Buzhen and Dai, Bo and Komura, Taku and Wang, Jingbo},
  booktitle={Proceedings of the Computer Vision and Pattern Recognition Conference},
  pages={5379--5391},
  year={2025}
}

@article{zhang2022motiondiffuse,
  title={MotionDiffuse: Text-Driven Human Motion Generation with Diffusion Model},
  author={Zhang, Mingyuan and Cai, Zhongang and Pan, Liang and Hong, Fangzhou and Guo, Xinying and Yang, Lei and Liu, Ziwei},
  journal={arXiv preprint arXiv:2208.15001},
  year={2022}
}

@inproceedings{petrovich22temos,
  title = {{TEMOS}: Generating diverse human motions from textual descriptions},
  author = {Petrovich, Mathis and Black, Michael J. and Varol, G{\"u}l},
  booktitle = {European Conference on Computer Vision ({ECCV})},
  year = {2022}
}

@InProceedings{Guo_2022_CVPR,
  author    = {Guo, Chuan and Zou, Shihao and Zuo, Xinxin and Wang, Sen and Ji, Wei and Li, Xingyu and Cheng, Li},
  title     = {Generating Diverse and Natural 3D Human Motions From Text},
  booktitle = {Proceedings of the IEEE/CVF Conference on Computer Vision and Pattern Recognition (CVPR)},
  month     = {June},
  year      = {2022},
  pages     = {5152-5161}
}

@article{humantomato,
  title={HumanTOMATO: Text-aligned Whole-body Motion Generation},
  author={Lu, Shunlin and Chen, Ling-Hao and Zeng, Ailing and Lin, Jing and Zhang, Ruimao and Zhang, Lei and Shum, Heung-Yeung},
  journal={arxiv:2310.12978},
  year={2023}
}

@article{liang2024intergen,
  title={Intergen: Diffusion-based multi-human motion generation under complex interactions},
  author={Liang, Han and Zhang, Wenqian and Li, Wenxuan and Yu, Jingyi and Xu, Lan},
  journal={International Journal of Computer Vision},
  pages={1--21},
  year={2024},
  publisher={Springer}
}

@inproceedings{tevet2023human,
  title={Human Motion Diffusion Model},
  author={Guy Tevet and Sigal Raab and Brian Gordon and Yoni Shafir and Daniel Cohen-or and Amit Haim Bermano},
  booktitle={The Eleventh International Conference on Learning Representations },
  year={2023},
  url={https://openreview.net/forum?id=SJ1kSyO2jwu}
}

@inproceedings{huang2024stablemofusion,
  title={Stablemofusion: Towards robust and efficient diffusion-based motion generation framework},
  author={Huang, Yiheng and Yang, Hui and Luo, Chuanchen and Wang, Yuxi and Xu, Shibiao and Zhang, Zhaoxiang and Zhang, Man and Peng, Junran},
  booktitle={Proceedings of the 32nd ACM International Conference on Multimedia},
  pages={224--232},
  year={2024}
}

@inproceedings{razali2023action,
  title={Action-conditioned generation of bimanual object manipulation sequences},
  author={Razali, Haziq and Demiris, Yiannis},
  booktitle={Proceedings of the AAAI conference on artificial intelligence},
  volume={37},
  number={2},
  pages={2146--2154},
  year={2023}
}

@article{li2023object,
  title={Object motion guided human motion synthesis},
  author={Li, Jiaman and Wu, Jiajun and Liu, C Karen},
  journal={ACM Transactions on Graphics (TOG)},
  volume={42},
  number={6},
  pages={1--11},
  year={2023},
  publisher={ACM New York, NY, USA}
}

@inproceedings{ghosh2023imos,
  title={IMoS: Intent-Driven Full-Body Motion Synthesis for Human-Object Interactions},
  author={Ghosh, Anindita and Dabral, Rishabh and Golyanik, Vladislav and Theobalt, Christian and Slusallek, Philipp},
  booktitle={Computer Graphics Forum},
  volume={42},
  number={2},
  pages={1--12},
  year={2023},
  organization={Wiley Online Library}
}

@inproceedings{li2024controllable,
  title={Controllable human-object interaction synthesis},
  author={Li, Jiaman and Clegg, Alexander and Mottaghi, Roozbeh and Wu, Jiajun and Puig, Xavier and Liu, C Karen},
  booktitle={European Conference on Computer Vision},
  pages={54--72},
  year={2024},
  organization={Springer}
}

@inproceedings{li2024task,
  title={Task-oriented human-object interactions generation with implicit neural representations},
  author={Li, Quanzhou and Wang, Jingbo and Loy, Chen Change and Dai, Bo},
  booktitle={Proceedings of the IEEE/CVF Winter Conference on Applications of Computer Vision},
  pages={3035--3044},
  year={2024}
}

@inproceedings{zeng2025chainhoi,
  title={Chainhoi: Joint-based kinematic chain modeling for human-object interaction generation},
  author={Zeng, Ling-An and Huang, Guohong and Wei, Yi-Lin and Gu, Shengbo and Tang, Yu-Ming and Meng, Jingke and Zheng, Wei-Shi},
  booktitle={Proceedings of the Computer Vision and Pattern Recognition Conference},
  pages={12358--12369},
  year={2025}
}

@inproceedings{wu2025human,
  title={Human-object interaction from human-level instructions},
  author={Wu, Zhen and Li, Jiaman and Xu, Pei and Liu, C Karen},
  booktitle={Proceedings of the IEEE/CVF International Conference on Computer Vision},
  pages={11176--11186},
  year={2025}
}

@inproceedings{hassan2023synthesizing,
  title={Synthesizing physical character-scene interactions},
  author={Hassan, Mohamed and Guo, Yunrong and Wang, Tingwu and Black, Michael and Fidler, Sanja and Peng, Xue Bin},
  booktitle={ACM SIGGRAPH 2023 Conference Proceedings},
  pages={1--9},
  year={2023}
}

@inproceedings{jia2024sceneverse,
  title={Sceneverse: Scaling 3d vision-language learning for grounded scene understanding},
  author={Jia, Baoxiong and Chen, Yixin and Yu, Huangyue and Wang, Yan and Niu, Xuesong and Liu, Tengyu and Li, Qing and Huang, Siyuan},
  booktitle={European Conference on Computer Vision},
  pages={289--310},
  year={2024},
  organization={Springer}
}

@article{daiya2026sketch2colab,
  title={Sketch2Colab: Sketch-Conditioned Multi-Human Animation via Controllable Flow Distillation},
  author={Daiya, Divyanshu and Bera, Aniket},
  journal={arXiv preprint arXiv:2603.02190},
  year={2026}
}

@article{tessler2024maskedmimic,
  title={Maskedmimic: Unified physics-based character control through masked motion inpainting},
  author={Tessler, Chen and Guo, Yunrong and Nabati, Ofir and Chechik, Gal and Peng, Xue Bin},
  journal={ACM Transactions On Graphics (TOG)},
  volume={43},
  number={6},
  pages={1--21},
  year={2024},
  publisher={ACM New York, NY, USA}
}

@article{xu2026interprior,
  title={InterPrior: Scaling Generative Control for Physics-Based Human-Object Interactions},
  author={Xu, Sirui and Schulter, Samuel and Ziyadi, Morteza and He, Xialin and Fei, Xiaohan and Wang, Yu-Xiong and Gui, Liangyan},
  journal={arXiv preprint arXiv:2602.06035},
  year={2026}
}

@article{wang2026unleashing,
  title={Unleashing Guidance Without Classifiers for Human-Object Interaction Animation},
  author={Wang, Ziyin and Xu, Sirui and Guo, Chuan and Zhou, Bing and Gong, Jiangshan and Wang, Jian and Wang, Yu-Xiong and Gui, Liang-Yan},
  journal={arXiv preprint arXiv:2603.25734},
  year={2026}
}

@inproceedings{tessler2025maskedmanipulator,
    author = {Tessler, Chen and Jiang, Yifeng and Coumans, Erwin and Luo, Zhengyi and Chechik, Gal and Peng, Xue Bin},
    title = {MaskedManipulator: Versatile Whole-Body Manipulation},
    year = {2025},
    booktitle={ACM SIGGRAPH Asia 2025 Conference Proceedings}
}

@inproceedings{peng2025hoi,
  title={Hoi-diff: Text-driven synthesis of 3d human-object interactions using diffusion models},
  author={Peng, Xiaogang and Xie, Yiming and Wu, Zizhao and Jampani, Varun and Sun, Deqing and Jiang, Huaizu},
  booktitle={Proceedings of the Computer Vision and Pattern Recognition Conference},
  pages={2878--2888},
  year={2025}
}

@misc{deepseekai2026deepseekv4highlyefficientmilliontoken,
  title={DeepSeek-V4: Towards Highly Efficient Million-Token Context Intelligence}, 
  author={DeepSeek-AI et.al.},
  year={2026},
  eprint={2606.19348},
  archivePrefix={arXiv},
  primaryClass={cs.CL},
  url={https://arxiv.org/abs/2606.19348}, 
}

@article{doi:10.1126/science.220.4598.671,
  author = {S. Kirkpatrick  and C. D. Gelatt  and M. P. Vecchi },
  title = {Optimization by Simulated Annealing},
  journal = {Science},
  volume = {220},
  number = {4598},
  pages = {671-680},
  year = {1983},
  doi = {10.1126/science.220.4598.671},
  URL = {https://www.science.org/doi/abs/10.1126/science.220.4598.671},
  eprint = {https://www.science.org/doi/pdf/10.1126/science.220.4598.671}
}

@misc{shan2024opendomaintextdrivensynthesis,
  title={Towards Open Domain Text-Driven Synthesis of Multi-Person Motions}, 
  author={Mengyi Shan and Lu Dong and Yutao Han and Yuan Yao and Tao Liu and Ifeoma Nwogu and Guo-Jun Qi and Mitch Hill},
  year={2024},
  eprint={2405.18483},
  archivePrefix={arXiv},
  primaryClass={cs.CV},
  url={https://arxiv.org/abs/2405.18483}, 
}

@misc{chang2025largescalemulticharacterinteractionsynthesis,
  title={Large-Scale Multi-Character Interaction Synthesis}, 
  author={Ziyi Chang and He Wang and George Alex Koulieris and Hubert P. H. Shum},
  year={2025},
  eprint={2505.14087},
  archivePrefix={arXiv},
  primaryClass={cs.GR},
  url={https://arxiv.org/abs/2505.14087}, 
}

@misc{ota2025pino,
  title = {PINO: Person-Interaction Noise Optimization for Long-Duration and Customizable Motion Generation of
  Arbitrary-Sized Groups},
  author = {Ota, Sakuya and Yu, Qing and Fujiwara, Kent and Ikehata, Satoshi and Sato, Ikuro},
  year = {2025},
  eprint = {2507.19292},
  archivePrefix = {arXiv},
  primaryClass = {cs.CV},
  url = {https://arxiv.org/abs/2507.19292}
}

@article{xie2023hierarchical,
  title={Hierarchical planning and control for box loco-manipulation},
  author={Xie, Zhaoming and Tseng, Jonathan and Starke, Sebastian and Van De Panne, Michiel and Liu, C Karen},
  journal={Proceedings of the ACM on Computer Graphics and Interactive Techniques},
  volume={6},
  number={3},
  pages={1--18},
  year={2023},
  publisher={ACM New York, NY, USA}
}

@inproceedings{pan2024synthesizing,
  title={Synthesizing physically plausible human motions in 3d scenes},
  author={Pan, Liang and Wang, Jingbo and Huang, Buzhen and Zhang, Junyu and Wang, Haofan and Tang, Xu and Wang, Yangang},
  booktitle={2024 International Conference on 3D Vision (3DV)},
  pages={1498--1507},
  year={2024},
  organization={IEEE}
}

@article{wu2024human,
  title={Human-object interaction from human-level instructions},
  author={Wu, Zhen and Li, Jiaman and Xu, Pei and Liu, C Karen},
  journal={arXiv preprint arXiv:2406.17840},
  year={2024}
}

@inproceedings{xiao2024unified,
  title={Unified human-scene interaction via prompted chain-of-contacts},
  author={Xiao, Zeqi and Wang, Tai and Wang, Jingbo and Cao, Jinkun and Zhang, Wenwei and Dai, Bo and Lin, Dahua and Pang, Jiangmiao},
  booktitle={International Conference on Learning Representations},
  volume={2024},
  pages={24450--24461},
  year={2024}
}

@article{deng2025human,
  title={Human-object interaction via automatically designed vlm-guided motion policy},
  author={Deng, Zekai and Shi, Ye and Ji, Kaiyang and Xu, Lan and Huang, Shaoli and Wang, Jingya},
  journal={arXiv preprint arXiv:2503.18349},
  year={2025}
}

@article{hart1968formal,
  title={A formal basis for the heuristic determination of minimum cost paths},
  author={Hart, Peter E and Nilsson, Nils J and Raphael, Bertram},
  journal={IEEE transactions on Systems Science and Cybernetics},
  volume={4},
  number={2},
  pages={100--107},
  year={1968},
  publisher={IEEE}
}

@article{dolgov2008practical,
  title={Practical search techniques in path planning for autonomous driving},
  author={Dolgov, Dmitri and Thrun, Sebastian and Montemerlo, Michael and Diebel, James},
  journal={ann arbor},
  volume={1001},
  number={48105},
  pages={18--80},
  year={2008}
}

@article{douglas1973algorithms,
  title={Algorithms for the reduction of the number of points required to represent a digitized line or its caricature},
  author={Douglas, David H and Peucker, Thomas K},
  journal={Cartographica: the international journal for geographic information and geovisualization},
  volume={10},
  number={2},
  pages={112--122},
  year={1973},
  publisher={University of Toronto Press}
}

\end{document}